\documentclass[12pt]{article}

\usepackage[margin=1in]{geometry}
\usepackage{amsmath,amssymb,amsthm,amsfonts}
\usepackage{bm}
\usepackage{graphicx}
\usepackage{booktabs}
\usepackage{algorithm}
\usepackage{algpseudocode}
\usepackage[round]{natbib}
\usepackage{setspace}
\usepackage[colorlinks=true,linkcolor=blue,citecolor=blue,urlcolor=blue]{hyperref}

\algrenewcommand{\algorithmicrequire}{\textbf{Input:}}
\algrenewcommand{\algorithmicensure}{\textbf{Output:}}

\makeatletter
\@ifundefined{appendices}{%
}{}
\makeatother

\graphicspath{{figure/}}

\newtheorem{theorem}{Theorem}

\newtheorem{definition}{Definition}
\newtheorem{assumption}{Assumption}

\def\wh{\widehat}

\newcommand{\mbN}{{\mathbb{N}}}

\def\calW{{\mathcal{W}}}

\newcommand{\calG}{{\mathcal G}}

\newcommand{\calX}{{\mathcal X}}

\newcommand{\mbR}{{\mathbb{R}}}

\def\cMMD{\mathrm{cMMD}}

\def\IcMMD{\mathrm{IcMMD}}
\def\OcMMD{\mathrm{OcMMD}}
\def\pover{\buildrel p\over\longrightarrow}
\newcommand{\E}{{\mathbb{E}}}

\DeclareMathOperator{\softmax}{softmax}

\allowdisplaybreaks

\begin{document}

\title{Optimal Mixture-of-Experts Model Averaging\\ for Conditional Generative Models}

\author{
Shijin Gong\textsuperscript{1}\thanks{The authors are listed in alphabetical order and recognized as co-first authors.}
\and
Baihua He\textsuperscript{1}
\and
Xinyu Zhang\textsuperscript{2,1}\thanks{Corresponding author: Xinyu Zhang. Email: \href{mailto:xinyu@amss.ac.cn}{xinyu@amss.ac.cn}.}
}

\date{\small
\textsuperscript{1}School of Management, University of Science and Technology of China\\
\textsuperscript{2}Academy of Mathematics and Systems Science, Chinese Academy of Sciences
}

\maketitle

\begin{abstract}
Conditional generative models have emerged as powerful tools for sampling from target conditional distributions, driving substantial advances across a wide range of scientific and applied domains. As these models proliferate, practitioners often face multiple plausible generators whose performance can vary with the task, data, or input condition. We propose an optimal model averaging framework for conditional generative models, allowing candidate generators to be combined even when they are accessible only through conditional samples without tractable densities. Specifically, we use a sample-based maximum mean discrepancy between conditional distributions, which first leads to a static model averaging method, \textsc{StaticMA}, assigning fixed weights to different candidates. In addition, we develop \textsc{MoEMA} (mixture-of-experts model averaging), an input-adaptive method that parameterizes covariate-dependent weights through a softmax neural-network gate. We establish in-sample and out-of-sample asymptotic optimality for the proposed methods, together with consistency of the estimated adaptive weight function under regularity conditions. The framework applies directly to Euclidean responses and extends to unstructured data by combining our formulation with  fixed representation maps. Across a broad set of simulations and real-data studies spanning tabular, image, and text modalities, \textsc{MoEMA} generally improves over competing baselines, demonstrating the effectiveness of our proposed methods.

\medskip
\noindent\textbf{Keywords:} conditional generative model, frequentist model averaging, mixture-of-experts, maximum mean discrepancy, asymptotic optimality
\end{abstract}
\setstretch{1.5}

\section{Introduction} \label{sec:intro}
Over the past five years, generative artificial intelligence has moved rapidly from a frontier research area into widespread practical use. Modern generative models can produce natural-language text and code \citep[e.g.,][]{openai2023gpt4}, synthesise images from textual or multimodal inputs \citep[e.g.,][]{rombach2022high}, and generate dynamic video content \citep[e.g.,][]{brooks2024video}.
These multimodal generative capabilities have expanded the role of AI from content creation to broader forms of human augmentation, supporting key activities such as writing, design, programming, and scientific discovery, while also enabling a wide range of applications across many other domains.

At the methodological core of this progress lies a statistical object that has been studied for decades: the conditional distribution of a response given an input. In generative AI, the input may be a prompt, a covariate vector, or other objects, and the goal is not merely to predict a single response, but to represent the range of plausible responses and their associated distribution. From a statistical perspective, two lines of work are especially relevant.
The first, with a long and well-developed history, models the response distribution explicitly, as in kernel density estimation \citep[e.g.,][]{silverman1986density}, conditional density estimation \citep[e.g.,][]{fan1996estimation}, and distributional regression \citep[e.g.,][]{hothorn2014conditional}.
The second strand is more recent and closely connected to modern generative modelling: it learns a conditional sampler from data, a procedure that takes an input and generates samples approximating the corresponding conditional distribution \citep[e.g.,][]{zhou2023deep}.

The rapid development of generative modelling has produced a diverse collection of conditional generative models, so practitioners often face several plausible generators and model choice itself becomes a source of uncertainty. Rather than selecting a single generator, a natural alternative is to aggregate candidate generators. Existing work in generative modelling supports this perspective, showing that leveraging multiple generative models or components can improve generation performance relative to relying on a single model \citep{balaji2022ediff,rezaei2025more}. Frequentist model averaging (FMA) provides a classical statistical framework for such aggregation under model selection uncertainty, assigning weights to candidate estimators under a risk criterion \citep[e.g.,][]{hansen2007least}. However, classical FMA is not directly tailored to conditional generative models, and adapting it to this setting calls for new methodological and theoretical development.

In this paper, we propose an optimal model averaging framework for conditional generative models, where the goal is to approximate a target conditional distribution and the candidate conditional distributions are accessible only through samples generated by the corresponding models. In this setting, classical criteria that require evaluating likelihoods, such as Kullback--Leibler (KL) divergence, are not fully feasible across candidates. We therefore apply an integral probability metric (IPM) \citep{muller1997integral} to conditional distributions. Choosing the function class in the IPM to be the unit ball of a reproducing kernel Hilbert space (RKHS), whose reproducing kernel is defined on the response space, yields the maximum mean discrepancy (MMD), a sample-based criterion estimable through kernel mean embeddings \citep{gretton2012kernel}. This leads to the proposed static model averaging method, \textsc{StaticMA}, which assigns weights on the simplex and, under the MMD objective, reduces to a quadratic programming problem.

\textsc{StaticMA} uses a single weight vector over the entire covariate space, thereby treating the relative importance of the candidate generators as the same for all inputs. This restriction can be strong for modern conditional generation. For example, large-scale evaluations show that no single language model dominates uniformly across tasks or domains \citep{bommasani2023helm}. Candidate generators may differ in architecture, training objective, or training data, and therefore may better approximate the target conditional distribution in different regions of the input space. To capture this heterogeneity, we introduce a covariate-dependent weight function, drawing on ideas from mixture-of-experts models \citep{jacobs1991adaptive}.
This gives the proposed input-adaptive method, \textsc{MoEMA} (mixture-of-experts model averaging), which parameterises the weight function by a softmax-gated neural network and is fitted using the same MMD-based objective. The framework accommodates generators that are trained as part of the procedure or adopted from pretrained sources. In either case, the aggregation step requires only the ability to generate conditional samples from each candidate model, and no generator needs to expose a tractable density.

On the theoretical side, the criterion for conditional distributions and the input-adaptive weight function introduce complications largely absent from classical optimal FMA, where the weights live in a finite-dimensional simplex and the criterion is a residual- or likelihood-based scalar risk. In this setting, we develop asymptotic theory for \textsc{MoEMA} and establish three results. The first two are asymptotic optimality results in the classical sense: both the in-sample risk (Theorem~\ref{thm:opt-adaptiveL}) and the out-of-sample risk (Theorem~\ref{thm:opt-adaptiveR}) of the method with data-driven weights are asymptotically equivalent to those of the infeasible oracle using the population-optimal weight function. The third is a weight-consistency result (Theorem~\ref{th:weight}): the estimated weight function itself converges to the population-optimal weight set, so the learned gate identifies which generators are locally relevant. 
The proofs address two complications that are not covered by existing FMA theory: the neural-network approximation error of a H\"older-class weight function and the dependence of the criterion on the fitted generators, the latter via an algorithmic-stability argument. Parallel results for \textsc{StaticMA} hold under strictly weaker conditions (Theorem~\ref{thm:opt-FL}-\ref{th:weightF}). To our knowledge, this provides the first model averaging theory for conditional generative models, and the first optimal model averaging theory built around a sample-based MMD criterion for comparing conditional distributions. Together with the input-adaptive extension realised by \textsc{MoEMA}, these results extend optimal FMA along three axes: from point prediction to conditional distribution estimation, from evaluable likelihoods to sample-based discrepancies, and from constant weights to input-adaptive weight functions.

This framework accommodates a broad range of tasks with different response types. For Euclidean responses, it applies directly with a kernel on the response space; for unstructured data, including images, text, or others, it extends through a feature-space variant (Section~\ref{sec:task-instances}) that composes the kernel with a fixed representation map and leaves the criterion otherwise unchanged.
We validate the methodology with extensive simulation and real-data studies spanning Euclidean, image, and text responses. Across these regimes, \textsc{MoEMA} improves over both the best single generator and other competing baselines.

The remainder of the paper is organised as follows. Section~\ref{sec:related} reviews related work. Section~\ref{sec:setup} formalises the problem and introduces the MMD criterion. Section~\ref{sec:method} develops the proposed \textsc{StaticMA} and \textsc{MoEMA} methods together with their feature-space extension to unstructured data. Section~\ref{sec:theory} establishes our theoretical results, Sections~\ref{sec:sim}--\ref{sec:real} report simulation and real-data studies, and Section~\ref{sec:disc} concludes.

\section{Related Work} \label{sec:related}

\subsection{Conditional Generative Modelling and Ensembles}
Conditional generative modelling aims to learn the distribution of an output given covariates, labels, prompts, or other contextual information. Existing approaches include conditional adversarial generators \citep{mirza2014conditional}, conditional variational autoencoders \citep{sohn2015learning}, normalising flows \citep{papamakarios2021flows}, diffusion and score-based methods for conditional sampling and forecasting \citep{ho2020denoising,song2021score}, and autoregressive language models that sample continuations conditional on prompts \citep{radford2019language,openai2023gpt4}. Other related lines include statistically grounded methods for conditional sampling and distributional regression \citep{zhou2023deep,song2025wasserstein}, as well as generators for tabular data \citep{kotelnikov2022tabddpm,chang2024conditionalfollmer,madhusudhanan2025tabresflow}.

Generator mixtures, ensembles, and model-composition methods provide another related perspective. Earlier work based on generative adversarial networks (GANs) used multiple generators to improve mode coverage \citep[e.g.,][]{ghosh2018multi}. More recent work combines pretrained generators at inference time, including diffusion composition through score- or energy-based operations \citep[e.g.,][]{du2023reduce} and decoding-time expert composition for controlled text generation \citep[e.g.,][]{liu2021dexperts}. Closest in spirit to model averaging, \citet{rezaei2025more} study global mixture optimisation for unconditional generative models. These approaches show that combining generators can improve diversity or performance, but many of them are tied to particular architectures. In contrast, our method treats each model as an arbitrary conditional sampler, uses only samples from them, allows input-adaptive weights, and is accompanied by theoretical guarantees.

\subsection{Frequentist Model Averaging}
Model averaging is a classical alternative to model selection, motivated by the instability and underreported uncertainty that can arise when inference proceeds as if a selected model were fixed in advance \citep{buckland1997model,claeskens2008model}. In FMA, candidate estimators or predictions are combined with data-dependent weights. Early and influential developments include focused and local-misspecification FMA \citep{hjort2003frequentist}, least-squares or Mallows model averaging \citep{hansen2007least}, and jackknife model averaging \citep{hansen2012jackknife}, among others.

Later work extends FMA beyond linear mean regression to richer responses and losses. Examples include model averaging for mixed-effects models \citep{zhang2014mixed}, Kullback-Leibler loss \citep{zhang2015kl}, generalised linear and generalised linear mixed-effects models \citep{zhang2016glm}, quantile regression \citep{lu2015quantile}, 
threshold models \citep{gao2019threshold}, nonlinear regression \citep{feng2022model} and time-varying or covariate-dependent weights \citep{sun2021timevarying,gao2026combining}. More general formulations allow broader candidate classes and broader loss functions, including K-fold cross-validation averaging \citep{zhang2023kfold} and unified general-loss criteria \citep{yu2025unified}. These developments are close to our work because they show that FMA can be adapted to richer losses and structures. Our contribution takes this direction further in a new way: we average conditional generators, and we choose weights by a sample-based conditional discrepancy between distributions. 

\subsection{Mixture-of-Experts}
Our input-adaptive method is related to mixture-of-experts methods, in which a gating function assigns inputs to specialised experts. Classical mixture-of-experts models introduced input-dependent routing among local predictors \citep{jacobs1991adaptive,jordan1994hierarchical}, and mixture density networks use neural networks to parameterise the mixing weights and components of conditional densities \citep{bishop1994mdn}. Recent statistical work has further studied inference and prediction sets for high-dimensional MoE regression models \citep{javanmard2025prediction}, as well as the use of MoE weighting to combine multiple prediction tools in prediction-powered inference \citep{gu2026prediction}. The same routing principle is also central to modern sparse MoE architectures for scaling deep networks, especially Transformer language models, where token-level gates activate a subset of internal experts to increase model capacity efficiently \citep[e.g.,][]{shazeer2017outrageously,dai2024deepseekmoe}; see \citet{cai2024survey} for a recent survey. \textsc{MoEMA} shares the idea of input-dependent weighting. However, the weighted objects are not internal components of a single architecture, but separate conditional generators, which can be of arbitrary types.


\section{Problem Setting and the MMD Criterion}
\label{sec:setup}

Let \(P(\cdot\mid x)\) denote the target conditional distribution on the response space \(\mathcal Y\) given \(x\in\mathcal X\). Depending on the task, the response \(y\) may be a scalar, a vector, an image, or another complex object. Data are observed as $\mathcal D_n=\bigl\{\bigl(x_i,\{y_{i,j}\}_{j=1}^{N_i}\bigr)\bigr\}_{i=1}^n$, where \(x_i\in\mathcal X\) is
a covariate and \(y_{i,1},\ldots,y_{i,N_i}\) are independent realisations from \(P(\cdot\mid x_i)\). Although many applications involve only a single response per covariate value, i.e., $N_i=1$, this notation also accommodates settings in which multiple responses are observed at the same covariate value.

For a positive integer \(N\), we define \([N]=\{1,\ldots,N\}\). For each \(m\in[M]\), let \(Q^{(m)}:\mathcal X \to \mathcal P(\mathcal Y)\) denote the conditional distribution induced by the \(m\)-th generative model, where \(\mathcal P(\mathcal Y)\) is the set of probability distributions on \(\mathcal Y\). The generative model is a sampler that, at each input \(x\in\mathcal X\), produces samples from \(Q^{(m)}(\cdot\mid x)\), which is specified up to trainable parameters (for example, the weights of a neural-network generative architecture), and \(\wh Q^{(m)}\) denotes the conditional distribution corresponding to parameters fitted on a training set \(\mathcal D^{(m)}\) (which may be different from \(\mathcal D_n\)) under some training algorithm and loss, or inherited from a pretrained model. Different generators in the pool may further differ in model class, architecture, or hyperparameters. We define the fitted pool as \(\widehat{\mathcal Q}=\{\wh Q^{(m)}\}_{m\in[M]}\), and treat each \(\wh Q^{(m)}\) as a black-box sampler accessed through conditional samples \(\wh Y^{(m)}(x)\sim\wh Q^{(m)}(\cdot\mid x)\), without requiring access to its conditional density. 
Section~D of the Supplementary Material reviews the main families of conditional generators and summarises whether each provides a tractable density.

To compare two conditional distributions on \(\mathcal Y\) using only samples, we use an integral probability metric (IPM). Let \(\mathcal F\) be a class of measurable functions on \(\mathcal Y\). For two conditional distributions \(P(\cdot\mid x)\) and \(Q(\cdot\mid x)\) on \(\mathcal Y\), the IPM between them is defined by
$$
d_{\mathcal F}\bigl(P(\cdot\mid x),Q(\cdot\mid x)\bigr)
=
\sup_{f\in\mathcal F}
\Bigl\{\Bigl|
\mathbb E_{Y\sim P(\cdot\mid x)}\bigl[f(Y)\bigr]
-
\mathbb E_{\widetilde{Y}\sim Q(\cdot\mid x)}\bigl[f(\widetilde Y)\bigr]\Bigr|
\Bigr\}.
$$
Different choices of the function class \(\mathcal F\) yield different IPMs. Our criterion is induced by a reproducing kernel Hilbert space \citep[RKHS;][]{aronszajn1950theory}, a choice that makes the resulting discrepancy estimable from samples through kernel evaluations alone. We first recall its definition.
\begin{definition}\label{def:rkhs}
Let \(\mathcal Y\) be a measurable space. A RKHS on \(\mathcal Y\) is a Hilbert space \(\mathcal H\) of real-valued functions on \(\mathcal Y\) such that there exists a positive-definite kernel \(k:\mathcal Y\times\mathcal Y\to\mathbb R\) satisfying: (i) \(k(\cdot,y)\in\mathcal H\) for every \(y\in\mathcal Y\); and (ii) \(f(y)=\langle f,k(\cdot,y)\rangle_{\mathcal H}\) for every \(f\in\mathcal H\) and \(y\in\mathcal Y\). The kernel \(k\) is called the reproducing kernel of \(\mathcal H\).
\end{definition}
We choose \(\mathcal F=\{f\in\mathcal H:\|f\|_{\mathcal H}\le 1\}\), where \(\mathcal H\) is an RKHS on \(\mathcal Y\) with kernel \(k:\mathcal Y\times\mathcal Y\to\mathbb R\). Under this choice, the IPM between the two conditional distributions at \(x\) specialises to the maximum mean discrepancy (MMD):
\begin{align*}
\mathrm{MMD}^2\bigl(P(\cdot\mid x),Q(\cdot\mid x)\bigr)
&=
\mathbb E_{Y,Y'\sim P(\cdot\mid x)}\!\bigl[k(Y,Y')\bigr]\\
&\quad-2\,\mathbb E_{Y\sim P(\cdot\mid x),\,\widetilde Y\sim Q(\cdot\mid x)}\!\bigl[k(Y,\widetilde Y)\bigr]
+\mathbb E_{\widetilde Y,\widetilde Y'\sim Q(\cdot\mid x)}\!\bigl[k(\widetilde Y,\widetilde Y')\bigr].
\end{align*}
This MMD measures discrepancy at each condition \(x\). To compare the two conditional distributions, we average it over the covariate distribution \(P_X\) and define conditional MMD (cMMD) as follows: for any two conditional distributions \(P\) and \(Q\) on \(\mathcal Y\),
\begin{equation}\label{eq:cmmd-def}
\mathrm{cMMD}^2(P,Q)
=\mathbb E_{X\sim P_X}\bigl[\mathrm{MMD}^2\bigl(P(\cdot\mid X),Q(\cdot\mid X)\bigr)\bigr].
\end{equation}
This definition is similar in spirit to averaged MMD of \citet{huang2022evaluating}. Note that \eqref{eq:cmmd-def} involves only expectations of kernel evaluations, so cMMD can be estimated from observed responses and generated conditional samples, making it usable without closed-form densities.
The technical conditions ensuring that cMMD is well defined and identifiable, in the sense that \(\mathrm{cMMD}(P,Q)=0\) entails \(P(\cdot\mid x)=Q(\cdot\mid x)\) for \(P_X\)-almost every \(x\), are provided in Section~A.2 of the Supplementary Material.

\section{Mixture-of-Experts Model Averaging Methodology}
\label{sec:method}

In this section, we derive the empirical cMMD objective for a weighted mixture of conditional generators, propose fixed-weight \textsc{StaticMA} and input-adaptive \textsc{MoEMA}, and extend the framework to unstructured responses through response and condition representations.

\subsection{Sample-Based cMMD Criterion}
We begin with the population mixture associated with a weight function \(w:\mathcal X\to\Delta^{M-1}\). For a generator pool \(\mathcal Q=\{Q^{(m)}\}_{m\in[M]}\), define
\[
Q(\cdot\mid x;w)=\sum_{m=1}^{M}w_m(x)Q^{(m)}(\cdot\mid x).
\]
Fix an arbitrary condition \(x\in\mathcal X\), the MMD between \(P(\cdot\mid x)\) and the mixture \(Q(\cdot\mid x;w)\) can be expressed as a function of the weights:
\begin{equation}
\begin{aligned}
  \mathrm{MMD}^{2}\bigl(P(\cdot\mid x),\,Q(\cdot\mid x;w)\bigr)
  &=
  A(x)
  -2\sum_{m=1}^{M}w_{m}(x)\,B_{m}(x)
  +\sum_{m=1}^{M}\sum_{m'=1}^{M}w_{m}(x)w_{m'}(x)\,C_{m,m'}(x), \notag
\end{aligned}
\end{equation}
where \(A(x)=\mathbb E_{Y,Y'\sim P(\cdot\mid x)}[k(Y,Y')]\), \(B_m(x)=\mathbb E_{Y\sim P(\cdot\mid x),\,\widetilde Y\sim Q^{(m)}(\cdot\mid x)}[k(Y,\widetilde Y)]\), and \(C_{m,m'}(x)=\mathbb E_{\widetilde Y\sim Q^{(m)}(\cdot\mid x),\,\widetilde Y'\sim Q^{(m')}(\cdot\mid x)}[k(\widetilde Y,\widetilde Y')]\).
Empirically, we approximate these terms by Monte Carlo samples. For any generator pool \(\mathcal Q\), we define its generated samples
$
\mathcal S_n(\mathcal Q)
=
\big\{
y_{i,\ell}^{(m)}(\mathcal Q)\sim Q^{(m)}(\cdot\mid x_i):
i\in[n],\ \ell\in[N_g],\ m\in[M]
\big\}
$
and the estimators obtained by $\mathcal S_n(\mathcal Q)$:
\begin{equation*}
\begin{aligned}
  \widehat B_{i,m}(\mathcal Q)
  &= \frac{1}{N_iN_g}\sum_{j=1}^{N_i}\sum_{\ell=1}^{N_g}
     k\bigl(y_{i,j},y_{i,\ell}^{(m)}(\mathcal Q)\bigr),\\
  \widehat C_{i,m,m'}(\mathcal Q)
  &=
  \begin{cases}
    \dfrac{1}{N_g(N_g-1)}
    \displaystyle\sum_{\ell=1}^{N_g}\sum_{\ell'\neq\ell}
    k\bigl(y_{i,\ell}^{(m)}(\mathcal Q),y_{i,\ell'}^{(m)}(\mathcal Q)\bigr), & m=m',\\[2.2ex]
    \dfrac{1}{N_g^{2}}
    \displaystyle\sum_{\ell=1}^{N_g}\sum_{\ell'=1}^{N_g}
    k\bigl(y_{i,\ell}^{(m)}(\mathcal Q),y_{i,\ell'}^{(m')}(\mathcal Q)\bigr), & m\neq m'.
  \end{cases}
\end{aligned}
\end{equation*}
For the diagonal terms \(m=m'\), the pairs with \(\ell=\ell'\) are excluded so that \(\widehat C_{i,m,m}(\mathcal Q)\) is an unbiased estimator of the corresponding \(C_{m,m}(x_i)\). Although \(A(x_i)\) cannot be estimated from a single response when \(N_i=1\), it does not depend on \(w\) and can therefore be omitted when estimating the weights. We define the empirical criterion
\begin{equation*}
\begin{aligned}
L_n(w,\mathcal Q)
&=-\frac{2}{n}\sum_{i=1}^{n}\sum_{m=1}^{M}w_{m}(x_i)\,\widehat B_{i,m}(\mathcal Q)+\frac{1}{n}\sum_{i=1}^{n}\sum_{m=1}^{M}\sum_{m'=1}^{M}
w_{m}(x_i)w_{m'}(x_i)\,\widehat C_{i,m,m'}(\mathcal Q).
\end{aligned}
\end{equation*}

\subsection{StaticMA and MoEMA}
We now introduce our proposed methods, starting with a fixed-weight version of the aggregation criterion. \textsc{StaticMA} restricts the weight to be constant over \(\mathcal X\), so that \(w_m(x)\equiv w_m\) for each \(m\). We estimate the fixed weight vector by solving the quadratic programming problem
\begin{equation*}
\widehat w
=
\arg\min_{w\in\Delta^{M-1}}
L_n(w,\widehat{\mathcal Q}).
\end{equation*}
This adopts the common model averaging perspective of combining candidate models through a weight vector optimised by the sample-based cMMD criterion. 

\textsc{MoEMA} allows the mixture weights to depend on the input. Let \(\mathcal W\) be a class of measurable functions \(w:\mathcal X\to\Delta^{M-1}\), where \(w(x)=(w_m(x))_{m\in[M]}^\top\), and let \(w_{\mathcal X_n}=(w(x_i))_{i\in[n]}^{\top}\) with \(\mathcal X_n=\{x_i\}_{i\in[n]}\). We model the weight function using a softmax gate, as in the MoE literature. Specifically, let \(g=(g_m)_{m\in[M]}^\top:\mathcal X\to\mathbb R^M\) be a score function satisfying the normalisation \(g_M(x)\equiv0\), and set
\begin{equation}\label{eq:softmax}
w_{m}(x)
=
\frac{\exp\{g_{m}(x)\}}
{\sum_{m'=1}^{M}\exp\{g_{m'}(x)\}},
\qquad m\in[M].
\end{equation}
Here, the normalisation \(g_M(x)\equiv0\) ensures the identifiability of the score function \(g\).
For estimation, we parameterise an unrestricted score function \(\widetilde g=(\widetilde g_m)_{m\in[M]}^\top:\mathcal X\to\mathbb R^M\) by a fully connected neural network:
\begin{align}\label{eq:g-mlp}
\widetilde g(x)
&=
E_D\sigma_{D-1}\{E_{D-1}\cdots\sigma_1(E_1x+b_1)\cdots+b_{D-1}\}+b_D, 
\end{align}
where \(E_l\) and \(b_l\) are network parameters, the hidden activations \(\sigma_l\) are ReLU functions, and the output dimension is \(M\). The normalised score is then obtained as \(g_m(x)=\widetilde g_m(x)-\widetilde g_M(x)\), \(m\in[M]\), which does not affect either the empirical criterion or the training implementation since \(w=\softmax(g)=\softmax(\widetilde g)\) . We denote by \(\mathcal W_n\) the resulting estimation class of softmax-gated networks of depth \(D\) and width \(W\). The weight function is then obtained by minimising the sample criterion directly over this class,
\begin{equation}\label{hatwab}
\widehat w
\in
\arg\min_{w\in\mathcal W_n}
L_n(w,\widehat{\mathcal Q}),
\end{equation}
and we write \(\widehat w_{\mathcal X_n}=(\widehat w(x_1),\ldots,\widehat w(x_n))^\top\) for its values at the observed covariates $\mathcal{X}_n$.
For a new input \(x\), prediction or generation proceeds by selecting a generator index \(\widehat m(x)\) with \(\mathbb P\{\widehat m(x)=m\}=\widehat w_m(x)\), \(m\in[M]\), and then sampling \(\wh Y\sim \wh Q^{(\widehat m(x))}(\cdot\mid x)\). Algorithm~1 in Section~E of the Supplementary Material summarises the full implementation.

In principle, the Monte Carlo size \(N_g\) can be any positive integer. Larger values of \(N_g\) reduce the estimator variance but increase the cost of computing the empirical kernel statistics. For each condition \(x_i\), the \(\widehat B\)-terms require \(O(MN_iN_g)\) kernel evaluations, while the \(\widehat C\)-terms require \(O(M^2N_g^2)\). In practice, \(N_g\) can therefore be chosen as large as computationally feasible. Other tuning parameters, including the kernel bandwidth and the neural-network hyperparameters in the adaptive gate, can be selected by cross-validation on the fitting criterion.

\subsection{Feature-Space Aggregation for Unstructured Data}
\label{sec:task-instances}
For image and text tasks, the response and the covariate may both be non-Euclidean. We use a response representation to compute discrepancies between conditional output distributions and a condition representation to evaluate the adaptive weight function.

For Euclidean responses, Algorithm~1 can be used directly with a kernel on \(\mathcal Y\). For unstructured responses, direct comparison in the raw response space is often not appropriate. We apply a pretrained representation model \(\psi:\mathcal Y\to\mathcal Z\) that transforms raw responses into Euclidean features. Then, we consider a kernel \(k_Z\) on \(\mathcal Z\), which induces the kernel \(k_{\psi}(y,y')=k_Z(\psi(y),\psi(y'))\), \(y,y'\in\mathcal Y\), on the original response space.
Consequently, the MMD computed with \(k_{\psi}\) can be expressed as 
\[
\begin{aligned}
\mathrm{MMD}_{k_{\psi}}^{2}\bigl(P(\cdot\mid x),&Q(\cdot\mid x)\bigr)
=\mathbb E_{Y,Y'\sim P(\cdot\mid x)}\bigl[k_Z(\psi(Y),\psi(Y'))\bigr]\\
&-2\,\mathbb E_{Y\sim P(\cdot\mid x),\,\widetilde Y\sim Q(\cdot\mid x)}\bigl[k_Z(\psi(Y),\psi(\widetilde Y))\bigr]
+\mathbb E_{\widetilde Y,\widetilde Y'\sim Q(\cdot\mid x)}\bigl[k_Z(\psi(\widetilde Y),\psi(\widetilde Y'))\bigr].
\end{aligned}
\]
The criterion therefore compares the conditional distributions of \(\psi(Y)\) and \(\psi(\widetilde Y)\), while generated samples remain elements of the original response space \(\mathcal Y\).

Next, we consider the covariate type. The gate function defined in \eqref{eq:g-mlp} requires \(x\) to be represented as a vector. When \(x\) is an unstructured input, we combine a pretrained representation model \(\varphi:\mathcal X\to\mathcal U\) with the same construction in \eqref{eq:softmax}, and parameterise the weight function by
\[
w_m(x)
=
\frac{\exp\{g_m(\varphi(x))\}}
{\sum_{r=1}^{M}\exp\{g_r(\varphi(x))\}},
\qquad m\in[M] .
\]
Here, taking \(\varphi(x)=x\) recovers the original version, and \textsc{StaticMA} is unchanged because its weights do not vary with \(x\). The generic implementation for such feature-space method is given in Algorithm~2 in Section~E of the Supplementary Material.

\section{Theoretical Results}
\label{sec:theory}
In this section, we establish three types of asymptotic guarantees for the proposed methods: in-sample optimality, out-of-sample optimality, and consistency of the estimated weights. We first analyse \textsc{StaticMA}, whose weight vector $w$ takes values in $\Delta^{M-1}$, and then \textsc{MoEMA}, whose weight function $w(\cdot)$ maps $\mathcal X$ into $\Delta^{M-1}$.

All asymptotic regimes considered in this paper correspond to \(n\to\infty\), whereas the within-block sample sizes \(N_i\) are allowed to be fixed. Define \(\underline N=\min_{1\le i\le n}N_i\). For a generator pool \(\mathcal Q\), define
\[
\widehat{\cMMD}^{2}(w,\mathcal D_n,\mathcal S_n(\mathcal Q))
=
A(\mathcal D_n)+L_n(w,\mathcal Q),
\qquad
A(\mathcal D_n)=\frac1n\sum_{i=1}^n A(x_i).
\]
Since \(A(\mathcal D_n)\) is independent of the weights, minimising $L_n(w,\mathcal Q)$ is equivalent to minimising \(\widehat{\cMMD}^{2}(\cdot,\mathcal D_n,\mathcal S_n(\widehat{\mathcal Q}))\) over the same admissible weight class. Here, for fixed weights method StaticMA, we have \(w_m(x_i)\equiv w_m\). Define \(\mathcal D'_n=\{(x'_i,\{y'_{i,j}\}_{j=1}^{N_i})\}_{i=1}^n\) as an independent copy of $\mathcal{D}_n$.
We use the same notational conventions for $\mathcal D'_n$ and define the corresponding generated sample set as
$
\mathcal S'_n(\mathcal Q)
=
\{
y_{i,\ell}^{\prime(m)}(\mathcal Q)\sim Q^{(m)}(\cdot\mid x'_i):
i\in[n],\ \ell\in[N_g],\ m\in[M]
\}.
$
Let \(\mathcal X'_n=\{x'_i\}_{i=1}^n\).
For any generator pool \(\mathcal Q\), define the in-sample and out-of-sample cMMD risks, respectively, by
\[
\begin{aligned}
\IcMMD^2(\cdot,\mathcal Q)
&=
\mathbb E_{\mathcal D_n}
\left\{
\mathbb E_{\mathcal S_n(\mathcal Q)}
\left[
\widehat{\cMMD}^{2}(\cdot,\mathcal D_n,\mathcal S_n(\mathcal Q))
\mid \mathcal D_n,\mathcal Q
\right]
\right\},\\
\OcMMD^2(\cdot,\mathcal Q)
&=
\mathbb E_{\mathcal D'_n,\mathcal S'_n(\mathcal Q)}
\left[
\widehat{\cMMD}^{2}(\cdot,\mathcal D'_n,\mathcal S'_n(\mathcal Q))
\mid \mathcal D_n,\mathcal Q
\right].
\end{aligned}
\]
Here, \(\widehat{\cMMD}^2(w,\mathcal D'_n,\mathcal S'_n(\mathcal Q))\) is computed by the same formula, with \((\mathcal D_n,\mathcal S_n(\mathcal Q))\) replaced by \((\mathcal D'_n,\mathcal S'_n(\mathcal Q))\). Note that when \(\mathcal Q=\widehat{\mathcal Q}\), the generators trained on \(\mathcal D_n\) are plugged into these definitions and held fixed inside the out-of-sample expectation. We first state the assumptions used in the analyses of both methods.

\begin{assumption}\label{convF}
For each $m\in[M]$, there exists a limiting conditional generator ${Q}^{(m)}_{\ast}$ such that
$$\max_{m\in[M]}\sup_{x\in\calX} \bigl[{\rm MMD}\bigl(\wh{Q}^{(m)}(\cdot\mid x),\;{Q}^{(m)}_{\ast}(\cdot\mid x)\bigr)\bigr]=O_p(a_n).$$
\end{assumption}
Assumption~\ref{convF} characterises the approximation properties of the candidate generators. A similar property has also been established in \citet{liang2021well}.
Throughout the remainder of this section, let $\mathcal Q_\ast$ denote the collection of limiting generators $\{Q_\ast^{(m)}\}_{m\in[M]}$. The corresponding limiting in-sample risk is then \(\IcMMD^2(\cdot,\mathcal Q_\ast)\), where the observed responses in \(\mathcal D_n\) remain distributed according to \(P\) but the generated samples in \(\mathcal S_n(\mathcal Q_\ast)\) are drawn from the limiting generator pool. The quantities \(\IcMMD^2(w,\mathcal Q_\ast)\) and \(\IcMMD^2(w_{\mathcal X_n},\mathcal Q_\ast)\) serve as the limiting risks for the static and input-adaptive analyses, respectively.

\begin{assumption}\label{kernelbd}
The kernel \(k\) on \(\mathcal Y\) is measurable, and there exists a finite constant \(K>0\) such that
$\sup_{y,y'\in\mathcal Y}|k(y,y')|\le K.
$
\end{assumption}
Assumption~\ref{kernelbd} is a mild kernel boundedness condition that ensures the per-condition MMD is well defined uniformly in $x$, and hence so is the cMMD criterion; it is satisfied with $K=1$ by any bounded kernel, including the Gaussian and Laplace kernels. In the remainder of this section, we study the asymptotic properties of our proposed methods.

\subsection{Asymptotic Properties for Static Model Averaging}

We first analyse \textsc{StaticMA}, whose weight vector $w$ takes values in $\Delta^{M-1}$. Let $\xi_n=\inf_{w\in\Delta^{M-1}}\IcMMD^2(w,\mathcal Q_\ast)$ denote the minimum limiting in-sample risk over the simplex. The following assumption couples $\xi_n$ with the number of candidate generators $M$ and the approximation rate $a_n$ from Assumption~\ref{convF}.

\begin{assumption}\label{DNNDWF}
$\max\{n^{-1/2},a_n, (\log M)^{1/2} N_g^{-1/2}\}/\xi_n=o(1)$,
$\sqrt{Mn^{-1}(\underline N^{-1}+N_g^{-1})\log\!\left({Kn}\right)}/{\xi_n}=o(1)$, and ${M\log\!\left({Kn}\right)}/(n{\xi_n})=o(1)$.
\end{assumption}
The first condition controls the empirical fluctuation and the generator approximation error, while the second and third conditions characterise the interplay among the number of candidate generators, the sample size, the generator sample size, and the complexity of the weight-function class relative to the oracle risk level \(\xi_n\). Similar conditions have also been imposed in \citet{yu2025unified}.
\begin{theorem}\label{thm:opt-FL}
Under Assumptions \ref{convF}--\ref{DNNDWF}, as $n\to\infty$,
$$\sup_{w\in \Delta^{M-1}}\left|\wh{\cMMD}^2(w,\mathcal D_n,\mathcal S_n(\widehat{\mathcal Q}))-\IcMMD^2(w,\mathcal Q_\ast)\right|\pover 0$$
and
\begin{align*}
\frac{\IcMMD^2(\wh{w},\mathcal Q_\ast)}{\inf_{w\in\Delta^{M-1}}\IcMMD^2(w,\mathcal Q_\ast)}\pover 1.
\end{align*}
\end{theorem}
Theorem~\ref{thm:opt-FL} shows that $\wh{\cMMD}^2(w,\mathcal D_n,\mathcal S_n(\widehat{\mathcal Q}))$ is a uniformly consistent estimator of the limiting risk $\IcMMD^2(w,\mathcal Q_\ast)$ and establishes the optimality of the proposed \textsc{StaticMA}, showing that its in-sample risk is asymptotically equivalent to that of the infeasible best model averaging method. The proof of Theorem~\ref{thm:opt-FL} is deferred to Subsection~C.1
in the Supplementary Material.

We next establish an analogous optimality result for the out-of-sample risk. In the out-of-sample setting, expectation is taken over a fresh sample $\mathcal D'_n$ while the fitted generators $\wh Q^{(m)}$ are held fixed at the values obtained from $\mathcal D_n$, so the analysis requires quantitative control on the sensitivity of each $\wh Q^{(m)}$ to its training data. This is provided by the uniform stability framework of \citet{bousquet2002stability}. We introduce the definition of uniform stability for learning the generators in the following. 
\begin{definition}
An algorithm $\mathcal{A}$ has uniform stability $\beta_n$ with respect to the loss function $\mathcal{T}$ if for all $\mathcal D_n=\{V_i\}_{i=1}^n,
V_i=(x_i,\{y_{i,j}\}_{j=1}^{N_i}),$ and all $i\in[n]$,
$$\left\| \mathcal{T}(V; \mathcal{A}_{\mathcal{D}_n}) - \mathcal{T}(V; \mathcal{A}_{\mathcal{D}_n^{\setminus i}}) \right\| \le \beta_n,$$
where $V$ is an independent copy of $\big\{(x_i,\{y_{i,j}\}_{j=1}^{N_i})\big\}$, $\mathcal{T}(V; \mathcal{A})$ denotes the loss function evaluated at the data point $V$, and $\mathcal{D}_n^{\setminus i}$ denotes the dataset with the $i$-th element removed.
\end{definition}
The parameter $\beta_n$ bounds the supremum-norm change in the learned generator when a single training point is removed; for standard regularised learning algorithms one has $\beta_n=O(n^{-1})$ \citep{bousquet2002stability}. Let $\xi'_n=\inf_{w\in\Delta^{M-1}} \OcMMD^2(w,\widehat{\mathcal Q})$. The next assumption couples this stability rate with $\xi'_n$, $n$, and $M$.
\begin{assumption}\label{DNNRF}
$n^{1/2}\beta_n/\xi'_n=o(1)$ and ${\beta_n (Mn\log (K n))^{1/2}}/{{\xi'_n}}=o(1)$, where $\beta_n$ is the {uniform-stability rate} for learning $\wh{Q}^{(m)}$ for each $m\in[M]$.
\end{assumption}

Assumption~\ref{DNNRF} imposes a comprehensive constraint on the stability-induced generalisation error to be asymptotically negligible relative to the oracle out-of-sample risk. 
In addition, Assumption~\ref{DNNRF} controls the accumulation of stochastic errors caused by the number of candidate generators and the complexity of the weight-function class.
\begin{theorem}\label{thm:opt-FR}
Under Assumptions \ref{convF}--\ref{DNNRF}, as $n\to\infty$,
\begin{align*}
\frac{\OcMMD^2(\wh{w},\widehat{\mathcal Q})}{\inf_{w\in\Delta^{M-1}}\OcMMD^2(w,\widehat{\mathcal Q})}\pover 1.
\end{align*}
\end{theorem}
Theorem~\ref{thm:opt-FR} is the out-of-sample counterpart of Theorem~\ref{thm:opt-FL}, which is of greater practical relevance, since out-of-sample performance is typically of primary interest. The proof is provided in Subsection~C.2 of the Supplementary Material.
We further show that the estimated weight vector $\wh w$ is itself consistent for the set of population minimisers of the limiting in-sample risk. Define \(\calW^\ast=\{w: w=\arg\min_{w\in\Delta^{M-1}}\IcMMD^2(w,\mathcal Q_\ast)\}\) and \(d(w,\calW^\ast)=\inf_{w'\in\calW^\ast}\|w-w'\|_2\).

\begin{theorem}\label{th:weightF}
Under Assumptions \ref{convF}--\ref{DNNDWF}, as $n\to\infty$,
$d(\wh{w},\calW^\ast)\pover 0.$
\end{theorem}
Theorem~\ref{th:weightF} shows the weak consistency of $\wh{w}$, in the sense that its distance to $\calW^\ast$ vanishes asymptotically. Similar results have been established by \citet{feng2022model}. However, our analysis is developed from a different perspective, as we explicitly account for the possible nonuniqueness of the optimal weight function \(w^\ast\). The proof of Theorem~\ref{th:weightF} is given in Subsection~C.3 of the Supplementary Material.
\subsection{Asymptotic Properties for Input-Adaptive Mixture-of-Experts Averaging}
We now analyse \textsc{MoEMA}, whose weight function is approximated by a softmax-gated neural network as in Section~\ref{sec:method}. The asymptotic analysis requires (i) a smoothness condition on the population-optimal weight function and (ii) a corresponding approximation-rate bound for the neural-network class. Both ingredients are encoded through a H\"older class, a standard smoothness class in nonparametric estimation.
Throughout this subsection, the score function \(g\) is the function defined in \eqref{eq:softmax}, with its \(M\)th coordinate fixed to zero for identifiability.
\begin{definition}
Let $\kappa=\alpha+\nu>0$, $\nu\in(0,1]$ and $\alpha=\lfloor\kappa\rfloor\in \mbN_0$, where $\lfloor\kappa\rfloor$ denotes the largest integer strictly smaller than $\kappa$ and $\mbN_0$ denotes the set of nonnegative integers. For a finite constant $c_0>0$, define the H\"{o}lder class $\calG^{\kappa}(\calX, c_0)$ as
\begin{align*}
    \calG^{\kappa}(\calX, c_0)=\left\{g:\calX\rightarrow \mbR: \max\limits_{\|a\|_1\leq \alpha} \|\partial^a g\|_{\infty}\leq c_0, \max\limits_{\|a\|_1=\alpha}\sup_{x\neq x^\prime} \frac{|\partial^a g(x)-\partial^a g(x^\prime)|}{\|x-x^\prime\|_2^{\nu}}\leq c_0\right\},
\end{align*}
where $\partial^a =\partial^{a_1}\cdots\partial^{a_q}$ with $a=(a_k)_{k\in[q]}^{\top} \in \mbN_0^q$, and $\|a\|_1=\sum_{k=1}^q a_k$.
\end{definition}
By definition, a function in $\calG^{\kappa}(\calX, c_0)$ is $\alpha$-times differentiable over $\calX$, with derivatives up to order $\alpha$ uniformly bounded by $c_0$, and its $\alpha$-th derivatives are H\"{o}lder continuous with exponent $\nu$. 
\begin{assumption}\label{hold}
For the score function \(g=(g_m)_{m\in[M]}^\top\), each coordinate \(g_m\) belongs to the H\"{o}lder class \(\calG^{\kappa}(\calX,c_0)\), and \(\max_{m\in[M]}\sup_{x\in\calX}|g_m(x)|\le B_g\) for some finite constant \(B_g\).
\end{assumption}
{Assumption \ref{hold} introduces a basic smoothness condition on the continuous function $g$, together with a uniform bound.} Although the smoothness parameter may not be directly inferred in practice, the smoothness condition is rather reasonable and general because it encompasses a large subset of continuous functions. For example, by conservatively setting $\kappa=1$, Assumption \ref{hold} implies that the function is Lipschitz continuous.
\begin{assumption}\label{conRieZ}
{The covariate support} $\calX\subset\mathbb R^q$ {is contained in a compact} \(p\){-dimensional Riemannian manifold isometrically embedded in} \(\mathbb R^q\){, with} \(p \le q\).
\end{assumption}

Assumption~\ref{conRieZ} assumes that $\mathcal X$ lies on a low-dimensional manifold. While this assumption is not required to establish the convergence rate of the weight function, it enables a faster convergence rate when the ambient dimension $q$ is large.  Denote $\calW=\left\{w(x)\ \mbox{lies in}\;\Delta^{M-1}: g_m(x)\in \calG^{\kappa}(\calX, c_0) \right\}$, and recall from Section~\ref{sec:method} the neural-network gate class $\calW_{n}=\bigl\{w(x)\; \mbox{lies in}\;\Delta^{M-1}: g_m(x)\in \calG_{n}(\calX, D, W) \bigr\}$, where $\calG_{n}$ is the neural network with depth $D$ and width $W$. We further denote $\calW_{\calX_n}=\{w(x_1),\ldots,w(x_n):w(x_i)\in\calW\}$ and $\calW_{n,\calX_n}=\{w(x_1),\ldots,w(x_n):w(x_i)\in\calW_{n}\}$. Define $\xi_{\calX_n}=\inf_{w_{\calX_n}\in\calW_{\calX_n}}\IcMMD^2(w_{\calX_n},\mathcal Q_\ast)$.
\begin{assumption}\label{DNNDW}
$\sqrt{M(\underline N^{-1}+N_g^{-1}){SD\log(S)\log(Kn M)}}/(n^{1/2}\xi_{\calX_n})=o(1)$, $M^2n^{-\frac{\kappa}{2\kappa+p\log q}}/\xi_{\calX_n}=o(1)$,
${{MSD\log(S)\log(Kn M)}/{(n\xi_{\calX_n})}}=o(1)$, and $\max\{n^{-1/2},a_n\}/\xi_{\calX_n}=o(1)$, where $S$ is the size of the deep neural network.
\end{assumption}
Assumption~\ref{DNNDW} pertains to the dimensionality of the covariates, the depth and size of the neural network, and the number of generators in relation to $n$ and $\xi_{\calX_n}$. Assumption~\ref{DNNDW} also imposes a condition on the generator convergence rate relative to $\xi_{\calX_n}$.
\begin{theorem}\label{thm:opt-adaptiveL}
Under Assumptions \ref{convF}--\ref{kernelbd} and \ref{hold}--\ref{DNNDW}, if the network width and depth are chosen as $W=114(\lfloor\kappa\rfloor+1)(p\log q)^{\lfloor\kappa\rfloor+1}$ and $D=21(\lfloor\kappa\rfloor+1)^2\lceil n^{(p\log q)/2(p\log q +2\kappa)}\log_2(8n^{(p\log q)/2(p\log q +2\kappa)})\rceil$, then as $n\to\infty$,
$$\sup_{w_{\calX_n}\in\calW_{n,\calX_n}}\left|\wh{\cMMD}^2(w_{\calX_n},\mathcal D_n,\mathcal S_n(\widehat{\mathcal Q}))-\IcMMD^2(w_{\calX_n},\mathcal Q_\ast)\right|\pover 0$$
and
\begin{align*}
\frac{\IcMMD^2(\wh{w}_{\calX_n},\mathcal Q_\ast)}{\inf_{w_{\calX_n}\in\calW_{\calX_n}}\IcMMD^2(w_{\calX_n},\mathcal Q_\ast)}\pover 1.
\end{align*}
\end{theorem}
Theorem~\ref{thm:opt-adaptiveL} establishes the optimality of the proposed \textsc{MoEMA}, showing that its in-sample risk is asymptotically equivalent to that of the infeasible best model averaging method. In contrast to prior work on optimal model averaging, our analysis explicitly incorporates the approximation error induced by using a neural network to approximate the true weight function in the H\"{o}lder class, thereby providing a more complete characterisation of its performance. The proof of Theorem~\ref{thm:opt-adaptiveL} is technically challenging due to the complicated structure of the empirical cMMD criterion. In particular, the adaptive weight function \(w(\cdot)\) enters the criterion in a highly complex manner. Moreover, the oracle risk is defined over an infinite-dimensional continuous function space, whereas the estimated weight function is restricted to a neural-network function class. Consequently, the analysis must simultaneously control the neural-network approximation error and the estimation error. 
We defer the proof to Subsection~C.4 of the Supplementary Material. Denote $\xi'_{\calX'_n}=\inf_{{w}_{\calX'_n}\in\calW_{\calX'_n}} \OcMMD^2({w}_{\calX'_n},\widehat{\mathcal Q})$.
\begin{assumption}\label{DNNR}
$n^{1/2}\beta_n/\xi'_{\calX'_n}=o(1)$ and $\frac{M SD n \beta_n^2\log(S) \log (K M n)}{(\xi'_{\calX'_n})^2}=o(1)$.
\end{assumption}
\citet{bousquet2002stability} also impose constraints on the stability of the learning algorithm. Assumption~\ref{DNNR} further imposes constraints on the dimensionality of the covariates, the depth and size of the neural network, and the number of generators in relation to $n$ and $\xi'_{\calX'_n}$.
\begin{theorem}\label{thm:opt-adaptiveR}
Under Assumptions \ref{convF}--\ref{kernelbd} and \ref{hold}--\ref{DNNR}, if the network width and depth are chosen as $W=114(\lfloor\kappa\rfloor+1)(p\log q)^{\lfloor\kappa\rfloor+1}$ and $D=21(\lfloor\kappa\rfloor+1)^2\lceil n^{(p\log q)/2(p\log q +2\kappa)}\log_2(8n^{(p\log q)/2(p\log q +2\kappa)})\rceil$, then as $n\to\infty$,
\begin{align*}
\frac{\OcMMD^2(\wh{w}_{\calX'_n},\widehat{\mathcal Q})}{\inf_{w_{\calX'_n}\in\calW_{\calX'_n}}\OcMMD^2(w_{\calX'_n},\widehat{\mathcal Q})}\pover 1.
\end{align*}
\end{theorem}
Theorem~\ref{thm:opt-adaptiveR} establishes the asymptotic out-of-sample optimality of the proposed adaptive model averaging procedure. Specifically, the out-of-sample risk achieved by the learned weight function asymptotically matches that of the infeasible oracle weight function within the entire neural-network weight class.
In parallel with Theorem~\ref{th:weightF}, we further show that the estimated weight function is consistent for the set of population minimisers of the limiting in-sample risk over $\calW_{\calX_n}$. Define
$\calW_{\calX_n}^\ast=\{w(x): w(x)=\arg\min_{w_{\calX_n}\in\calW_{\calX_n}}\IcMMD^2(w_{\calX_n},\mathcal Q_\ast)\}$ and $d(w_{\calX_n},\calW_{\calX_n}^\ast)=\inf_{w'_{\calX_n}\in\calW_{\calX_n}^\ast}\E_{\calX_n}\bigl[\|w_{\calX_n}-w'_{\calX_n}\|_2\bigr]=\sum_{i=1}^n \inf_{w'_{\calX_n}\in\calW_{\calX_n}^\ast}\E_{x_i}\bigl[\|w(x_i)-w'(x_i)\|_2\bigr]/n.$

\begin{theorem}\label{th:weight}
Under Assumptions \ref{convF}--\ref{kernelbd} and \ref{hold}--\ref{DNNDW}, if the network width and depth are chosen as $W=114(\lfloor\kappa\rfloor+1)(p\log q)^{\lfloor\kappa\rfloor+1}$ and $D=21(\lfloor\kappa\rfloor+1)^2\lceil n^{(p\log q)/2(p\log q +2\kappa)}\log_2(8n^{(p\log q)/2(p\log q +2\kappa)})\rceil$, then as $n\to\infty$,
$d(\wh{w}_{\calX_n},\calW_{\calX_n}^\ast)\pover 0.$
\end{theorem}

Theorem~\ref{th:weight} establishes the consistency of the learned adaptive weight function. Specifically, the weight vector \(\widehat w_{\calX_n}\) obtained from \eqref{hatwab} converges to the optimal weight set \(\mathcal W^\ast_{\calX_n}\), in the sense that its distance to \(\mathcal W^\ast_{\calX_n}\) vanishes asymptotically. This result is substantially stronger than risk consistency, as it guarantees that the learned adaptive weights themselves recover the oracle weighting structure asymptotically. The proof of Theorem~\ref{th:weight} is given in Subsection~C.6 of the Supplementary Material.


\section{Simulation Studies}\label{sec:sim}
We conduct two simulation designs, one varying the model family with a fixed training sample and the other varying the training covariate distribution with a fixed model family. We compare \textsc{BestSingle}, \textsc{SimpleAvg} (the equal-weight baseline), \textsc{StaticMA}, and \textsc{MoEMA}; \textsc{BestSingle} denotes the best individual generator on test set for each metric and is used as an oracle per-metric model-selection benchmark. We further verify the weight-consistency result of Theorem~\ref{th:weight} on a separate experiment; see Section~F.2 of the Supplementary Material.

Both designs use the same structured conditional Gaussian mixture. Let \(K_{\rm mix}=12\), and let
\(\Theta=\{(\mu_{xk},\mu_{yk},A_k,\sigma_{yk})\}_{k=1}^{K_{\rm mix}}\)
be a fixed set of component parameters, drawn once for each dimension
and then held fixed. The components are chosen to induce heterogeneous local conditional distributions across the covariate space.
Conditional on a latent component \(C\sim\mathrm{Unif}([K_{\rm mix}])\),
\[
X\mid C=k\sim
\mathcal N(\mu_{xk},\sigma_X^2I_{d_x}),
\qquad
Y\mid X=x,C=k\sim
\mathcal N\{\mu_{yk}+A_k(x-\mu_{xk}),\sigma_{yk}^2I_{d_y}\}.
\]
In other words, $P(y|x)=\sum_{k=1}^{K_{\rm mix}} \pi_k(x)\,
\phi_{d_y}\!\{y;\mu_{yk}+A_k(x-\mu_{xk}),\sigma_{yk}^2I_{d_y}\}$,
where \(\phi_d(\cdot;\mu,\Sigma)\) denotes the \(d\)-dimensional Gaussian density and
$
\pi_k(x)=
{\phi_{d_x}(x;\mu_{xk},\sigma_X^2I_{d_x})}/
{\sum_{\ell=1}^{K_{\rm mix}}\phi_{d_x}(x;\mu_{x\ell},\sigma_X^2I_{d_x})}.
$
We consider \((d_x,d_y)\in\{(1,1),(5,1),(10,1),(5,3)\}\), vary \(n\in\{250,500,1000,2000\}\), and evaluate all methods on an independent test sample of size \(1000\). The construction of the fixed parameter set \(\Theta\) is given in Section~F.3 of the Supplementary Material.

\begin{figure}[htbp]
\centering
\includegraphics[width=\linewidth]{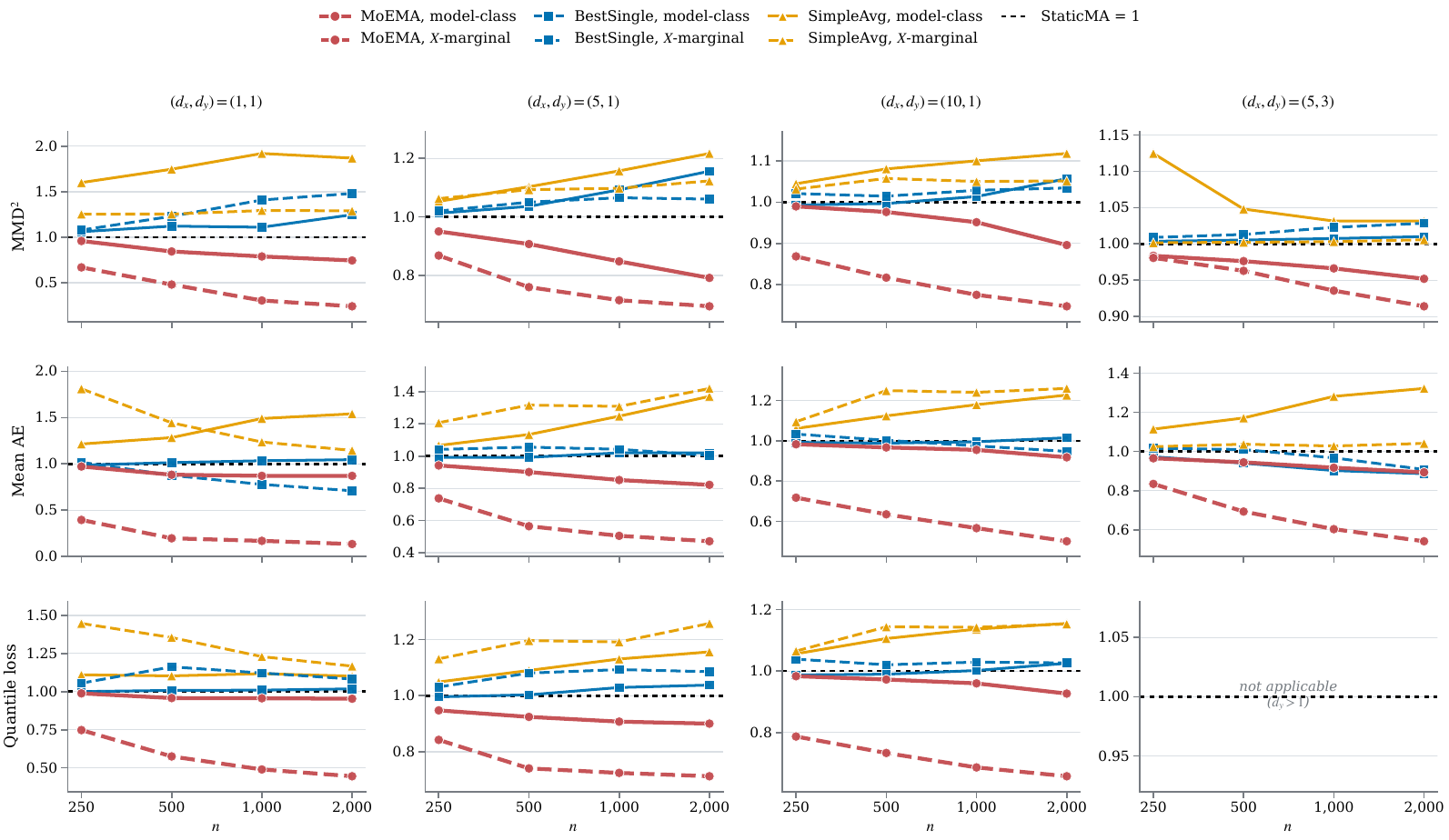}
\caption{Simulation results across sample sizes. Rows report \(\mathrm{MMD}^2\), MeanAE, and QuantileLoss; columns show dimension settings. Values are normalised by \textsc{StaticMA}; solid/dashed lines denote architecture/training-data heterogeneity.}
\label{fig:simulation_multidimensional}
\end{figure}

\subsection{Heterogeneous Architectures}

In the first design, the candidate pool consists of four conditional generators from variational autoencoder, diffusion, and normalising-flow families, with architecture or hyperparameter differences. All candidates are trained on the same observed sample \(\mathcal D_n\), on which the static and adaptive weights are also fit. We report three metrics: sample \(\mathrm{MMD}^2\) measures the full conditional distribution \(P(\cdot\mid x)\), while QuantileLoss and MeanAE measure the accuracy of conditional quantiles and conditional means computed from the generated samples. Details of the simulation settings and metric definitions are given in Section~F of the Supplementary Material. 

The results are shown by the solid curves in Figure~\ref{fig:simulation_multidimensional}, where each curve reports the ratio of a method's metric value to that of \textsc{StaticMA} under the same setting. The aggregation methods generally improve over the per-metric best single generator and equal-weight baseline, and \textsc{MoEMA} further improves on \textsc{StaticMA} throughout the four dimension settings. Relative reductions range from \(1.0\%\) to \(25.5\%\) for \(\mathrm{MMD}^2\), from \(1.7\%\) to \(17.8\%\) for MeanAE, and from \(1.1\%\) to \(10.0\%\) for QuantileLoss. See Table~S1 for the full numerical results.

\subsection{Heterogeneous Training Data}
In the second design we use the same DGP as in the first design but keep the model family fixed and vary only the training covariate distribution. The candidate pool consists of three diffusion-based generative regressors with the same architecture and optimisation settings. With \(u(x)=d_x^{-1/2}\sum_{j=1}^{d_x}x_j\), the three candidates are trained on samples from the original DGP conditional on \(u(X)<-1\), \(|u(X)|\leq 1\), and \(u(X)>1\), respectively. The dataset \(\mathcal D_n\) used to fit \textsc{StaticMA} and \textsc{MoEMA} is drawn from the original DGP. 

The dashed curves in Figure~\ref{fig:simulation_multidimensional} show that \textsc{MoEMA} continues to outperform the non-adaptive baselines. Relative to the second best method \textsc{StaticMA}, \textsc{MoEMA} reduces \(\mathrm{MMD}^2\), MeanAE, and QuantileLoss by \(1.9\%\)--\(75.9\%\), \(16.5\%\)--\(86.5\%\), and \(15.8\%\)--\(55.5\%\), respectively. These gains are substantially larger than those in the first design. The input-adaptive gate selects the locally relevant specialist for each input, which a constant weight vector cannot achieve. See Table~S2 for the full numerical results.

\section{Real-Data Studies}\label{sec:real}

We evaluate the proposed methods on real-world tasks spanning several response modalities and generator families. We begin with tabular probabilistic regression on two scalar-response benchmarks, then turn to image generation and inpainting, and finally consider a short-text continuation task with autoregressive language models. In all real-data tables, for each baseline cell value \(b\), \(\Delta=100\times(b-\textsc{MoEMA})/b\) reports the relative error reduction achieved by \textsc{MoEMA}, where positive values indicate lower error under \textsc{MoEMA}.

\subsection{Tabular Studies}
In this experiment, the condition is a covariate vector and we learn the conditional distribution of a scalar response. We use two datasets from the UCI Machine Learning Repository, \textsc{CT Slices}\footnote{\url{https://archive.ics.uci.edu/dataset/206/relative+location+of+ct+slices+on+axial+axis}} and \textsc{Protein}\footnote{\url{https://archive.ics.uci.edu/dataset/265/physicochemical+properties+of+protein+tertiary+structure}}. As in the simulation study, we consider two sources of generator diversity. The first varies the generator family while using a common training split: we combine a diffusion-based model \citep{han2022card}, a conditional F\"ollmer-flow model \citep{chang2024conditionalfollmer}, a diffusion-boosted tree model \citep{han2024dbt}, and the spline-flow model TabResFlow \citep{madhusudhanan2025tabresflow}, all trained on the full dataset. The second keeps the generator family fixed and varies the training covariate distribution: the candidate models are TabResFlow generators trained on different subsets, including one random subset \(\mathcal D_n\) and three subsets with different marginal distributions. Full details are provided in Section~G.1 of the Supplementary Material.

\begin{table}[htbp]
\centering
\small
\setlength{\tabcolsep}{3pt}
\renewcommand{\arraystretch}{0.94}
\begin{tabular}{llcccccc}
\toprule
\textbf{Dataset} & \textbf{Model} & \textbf{RMSE}$\downarrow$ & $\Delta$ & \textbf{Energy}$\downarrow$ & $\Delta$ & \textbf{Pinball}$\downarrow$ & $\Delta$ \\
\midrule
\multicolumn{8}{@{}l}{\textit{Panel A: Different generator families, common full split}} \\
\midrule
\textsc{CT Slices} & BestSingle & 1.776 & $+13.1\%$ & 1.421 & $+10.6\%$ & 0.318 & $+9.3\%$ \\
& SimpleAvg  & 2.177 & $+29.1\%$ & 2.357 & $+46.1\%$ & 0.591 & $+51.2\%$ \\
& StaticMA   & 1.703 & $+9.3\%$ & 1.408 & $+9.8\%$ & 0.316 & $+8.8\%$ \\
& \textbf{MoEMA} & \textbf{1.544} & -- & \textbf{1.270} & -- & \textbf{0.289} & -- \\
\midrule
\textsc{Protein} & BestSingle & 4.476 & $+0.0\%$ & 4.523 & $+0.0\%$ & 0.998 & $-0.0\%$ \\
& SimpleAvg  & 4.513 & $+0.8\%$ & 4.624 & $+2.2\%$ & 1.022 & $+2.3\%$ \\
& StaticMA   & 4.476 & $+0.0\%$ & 4.523 & $+0.0\%$ & 0.998 & $-0.0\%$ \\
& \textbf{MoEMA} & \textbf{4.476} & -- & \textbf{4.523} & -- & \textbf{0.999} & -- \\
\midrule
\multicolumn{8}{@{}l}{\textit{Panel B: TabResFlow generators, different training supports}} \\
\midrule
\textsc{CT Slices} & BestSingle & 5.448 & $+42.7\%$ & 4.743 & $+56.3\%$ & 1.072 & $+55.1\%$ \\
& SimpleAvg  & 9.301 & $+66.5\%$ & 6.417 & $+67.7\%$ & 1.518 & $+68.3\%$ \\
& StaticMA   & 9.329 & $+66.6\%$ & 6.255 & $+66.8\%$ & 1.481 & $+67.5\%$ \\
& \textbf{MoEMA} & \textbf{3.121} & -- & \textbf{2.074} & -- & \textbf{0.481} & -- \\
\midrule
\textsc{Protein} & BestSingle & 5.170 & $+15.0\%$ & 5.389 & $+21.4\%$ & 1.179 & $+20.4\%$ \\
& SimpleAvg  & 4.980 & $+11.8\%$ & 5.077 & $+16.6\%$ & 1.112 & $+15.6\%$ \\
& StaticMA   & 4.914 & $+10.6\%$ & 4.994 & $+15.2\%$ & 1.101 & $+14.7\%$ \\
& \textbf{MoEMA} & \textbf{4.393} & -- & \textbf{4.236} & -- & \textbf{0.939} & -- \\
\bottomrule
\end{tabular}
\renewcommand{\arraystretch}{1.0}
\caption{Tabular real-data experiments comparing aggregation rules across generator pools. Reported values are averages over five random repeats.}
\label{tab:real_reg_main}
\end{table}

Table~\ref{tab:real_reg_main} shows different patterns across the two scenarios. In Panel A, \textsc{StaticMA} improves over the best single generator on \textsc{CT Slices}, and \textsc{MoEMA} further reduces the reported metrics by \(8.8\%\)--\(51.2\%\) relative to the other methods. On \textsc{Protein}, the candidate generators provide less complementary information, and the learned mixtures perform similarly to the strongest individual generator. In Panel B, the role of input-dependent weighting is more pronounced. On \textsc{CT Slices}, the generators perform differently on different region subsets, and \textsc{MoEMA} improves across metrics by \(42.7\%\)--\(68.3\%\) relative to the other methods. On \textsc{Protein}, \textsc{StaticMA} improves over the \textsc{BestSingle} and \textsc{SimpleAvg}, and \textsc{MoEMA} further reduces the reported metrics by \(10.6\%\)--\(21.4\%\) relative to the other methods. Overall, \textsc{MoEMA} is the strongest method across different scenarios.

\subsection{Image Studies}
We conduct three image studies, including handwritten digits generation on MNIST\footnote{\url{https://docs.pytorch.org/vision/stable/generated/torchvision.datasets.MNIST.html}}, natural-image generation on ImageNet-100\footnote{\url{https://docs.pytorch.org/vision/stable/generated/torchvision.datasets.ImageNet.html}}, and masked-image inpainting on MNIST and CIFAR-10\footnote{\url{https://docs.pytorch.org/vision/stable/generated/torchvision.datasets.CIFAR10.html}}. Since the responses are images, these experiments use the feature-space version introduced in Section~\ref{sec:task-instances}. We evaluate generated images in a fixed feature space using Fr\'echet Inception Distance (FID) and Kernel Inception Distance (KID), two commonly used metrics for image generative models. See Section~G.2 of the Supplementary Material for mathematical formulations and implementation details.

\subsubsection{Class-Conditional Generation on MNIST}

We first consider class-conditional generation on MNIST, using the digit class as the conditioning variable. The candidate pool consists of three conditional diffusion generators \citep{ho2020denoising} trained on the same handwritten-digit data with different class distributions. The static and adaptive mixtures are fitted from MNIST training images grouped by digit class, and final performance is evaluated on the MNIST test set. In \textsc{MoEMA}, the gate maps the digit label to mixture weights. Table~\ref{tab:image_generation} shows that \textsc{MoEMA} improves both criteria over every baseline by a wide margin, with relative reductions ranging from \(60\%\) to over \(90\%\).

\subsubsection{Prompt-Conditional Generation on ImageNet-100}
We next consider image generation task on ImageNet-100, conditional on class names. The candidate pool is a heterogeneous collection of pretrained conditional generators from prior work \citep[e.g.,][]{dhariwal2021diffusion,brock2019large}. We fit the methods on ImageNet-100 training images grouped by class and evaluate them on a disjoint set of real images. The \textsc{MoEMA} uses a Contrastive Language--Image Pretraining (CLIP) text encoder \citep{radford2021learning} as the condition representation model \(\varphi\) in Section~\ref{sec:task-instances}.

\begin{table}[htbp]
\centering
\small
\setlength{\tabcolsep}{2.5pt}
\renewcommand{\arraystretch}{0.93}
\begin{tabular}{lcccc|cccc}
\toprule
& \multicolumn{4}{c|}{\textbf{MNIST}} & \multicolumn{4}{c}{\textbf{ImageNet-100}}\\
\cmidrule(lr){2-5}\cmidrule(lr){6-9}
\textbf{Method} & FID$\downarrow$ & $\Delta$ & KID ($\times 10^{-4}$)$\downarrow$ & $\Delta$ & FID$\downarrow$ & $\Delta$ & KID ($\times 10^{-3}$)$\downarrow$ & $\Delta$ \\
\midrule
BestSingle     & 0.018 & $+75.3\%$ & 1.001 & $+94.7\%$ & 13.021 & $+9.8\%$ & 1.964 & $+31.9\%$ \\
SimpleAvg      & 0.011 & $+60.1\%$ & 0.471 & $+88.7\%$ & 17.889 & $+34.3\%$ & 4.871 & $+72.5\%$ \\
StaticMA       & 0.016 & $+71.1\%$ & 0.783 & $+93.2\%$ & 11.802 & $+0.5\%$ & \textbf{1.317} & $-1.6\%$ \\
\textbf{MoEMA} & \textbf{0.004} & -- & \textbf{0.053} & -- & \textbf{11.747} & -- & 1.338 & -- \\
\bottomrule
\end{tabular}
\renewcommand{\arraystretch}{1.0}
\caption{Conditional image generation on MNIST and ImageNet-100 evaluated by FID and KID. Reported values are averages over five random repeats.}
\label{tab:image_generation}
\end{table}

On ImageNet-100, Table~\ref{tab:image_generation} shows that simple averaging fails to improve on the best single generator. \textsc{StaticMA} obtains most of the available improvement (about 10\% on FID and 32\% on KID relative to the per-metric best single generator) and performs similarly to \textsc{MoEMA}. 

\subsubsection{Masked-Image Inpainting on MNIST and CIFAR-10}
We then consider masked-image inpainting, with the centrally masked image as the condition and the completed image as the response. We use handwritten digits on MNIST and natural images on CIFAR-10, two settings in which a single partially observed input can have several plausible completions. For each dataset, we use five DDPMs \citep{ho2020denoising} from the same architecture family, trained under different subsets so that each candidate is exposed more heavily to a different subset of image classes. The static and adaptive mixtures are fitted on a random subset of the original dataset images and evaluated on held-out test images. 

\begin{table}[htbp]
\centering
\small
\setlength{\tabcolsep}{2.5pt}
\renewcommand{\arraystretch}{0.93}
\begin{tabular}{lcccc|cccc}
\toprule
& \multicolumn{4}{c|}{\textbf{MNIST}} & \multicolumn{4}{c}{\textbf{CIFAR-10}}\\
\cmidrule(lr){2-5}\cmidrule(lr){6-9}
\textbf{Method} & FID$\downarrow$ & $\Delta$ & KID ($\times 10^{-3}$)$\downarrow$ & $\Delta$ & FID$\downarrow$ & $\Delta$ & KID ($\times 10^{-2}$)$\downarrow$ & $\Delta$ \\
\midrule
BestSingle     & 0.192 & $+73.1\%$ & 2.152 & $+82.0\%$ & 1.291 & $+12.7\%$ & 3.222 & $+21.0\%$ \\
SimpleAvg      & 0.062 & $+16.4\%$ & 0.531 & $+27.1\%$ & 1.351 & $+16.5\%$ & 3.345 & $+23.9\%$ \\
StaticMA       & 0.059 & $+12.4\%$ & 0.388 & $+0.1\%$ & 1.323 & $+14.8\%$ & 3.360 & $+24.2\%$ \\
\textbf{MoEMA} & \textbf{0.052} & -- & \textbf{0.387} & -- & \textbf{1.128} & -- & \textbf{2.545} & -- \\
\bottomrule
\end{tabular}
\renewcommand{\arraystretch}{1.0}
\caption{Masked-image inpainting on MNIST and CIFAR-10 evaluated by FID and KID. Reported values are averages over five random repeats.}
\label{tab:inpainting}
\end{table}

Table~\ref{tab:inpainting} shows that on both datasets the gain reflects input-adaptive weighting rather than the selection of a single dominant generator: both \textsc{StaticMA} and \textsc{MoEMA} outperform the per-metric best individual DDPM, with relative reductions of roughly 14\%--78\% across datasets and metrics. Among the averaging rules, \textsc{MoEMA} improves over \textsc{StaticMA} by about 10\%--13\% on FID across both datasets and by roughly a quarter on CIFAR-10 KID, while on MNIST KID the two adaptive rules are essentially tied.

\subsection{Text Continuation Study}
We consider short-text continuation, where a generator produces a 24-token continuation from a 16-token prompt. We use four source domains from Hugging Face Datasets: \textsc{AG News}, \textsc{IMDb}, \textsc{DBPedia-14}, and \textsc{Yahoo Answers Topics}. We fine-tune four GPT-2 language models \citep{radford2019language}, one per source domain. These four generators are then aggregated with our proposed methods.

Although token-level likelihoods are available, the static and adaptive weights are fitted on a balanced held-out prompt-continuation set using the same sample-based MMD objective in a frozen GPT-2 feature space. After fitting, each method is evaluated on held-out prompts from the mixture of the four domains by comparing its continuation distribution with the observed continuations through MMD\(^2\) and energy distance in the same feature space. We also report the perplexity (PPL) induced by the fitted weights and the generators' token probabilities. Implementation details are given in Section~G.3 of the Supplementary Material.

\begin{table}[htbp]
\centering
\small
\setlength{\tabcolsep}{4pt}
\begin{tabular}{lcccccc}
\toprule
\textbf{Method} & MMD$^2\downarrow$ & $\Delta$ & Energy$\downarrow$ & $\Delta$ & PPL$\downarrow$ & $\Delta$ \\
\midrule
BestSingle     & 0.432 & $+6.8\%$ & 0.082 & $+9.5\%$ & 77.606 & $+39.4\%$ \\
SimpleAvg      & 0.413 & $+2.5\%$ & 0.077 & $+4.0\%$ & 47.681 & $+1.4\%$ \\
StaticMA       & 0.412 & $+2.3\%$ & 0.077 & $+3.7\%$ & 47.809 & $+1.7\%$ \\
\textbf{MoEMA} & \textbf{0.402} & -- & \textbf{0.074} & -- & \textbf{47.004} & -- \\
\bottomrule
\end{tabular}
\caption{Multi-domain short-continuation text generation evaluated in GPT-2 feature space. Reported values are averages over five random repeats.}
\label{tab:text_llm_main}
\end{table}

Table~\ref{tab:text_llm_main} shows that every averaging rule outperforms the per-metric best single generator by a wide margin, about 6\%--9\% on MMD\(^2\) and Energy and roughly 40\% on perplexity. On this comparison \textsc{MoEMA} improves over \textsc{StaticMA} on all three metrics by roughly 2\%--5\%, showing that the same sample-based aggregation mechanism extends to autoregressive language models without changing the generator training objective or relying on likelihood-based weight fitting.

\section{Discussion}\label{sec:disc}

This paper studies the optimal model averaging under conditional generative models. The framework defines aggregation through a sample-based MMD between conditional distributions, and proposes both fixed-weight \textsc{StaticMA} and input-adaptive \textsc{MoEMA}, the latter parameterised by a covariate-dependent weight function. The two methods are shown to be asymptotically optimal in both in-sample and out-of-sample risk, and the input-adaptive weight function is consistent for the optimal gate. Empirically, \textsc{StaticMA} and \textsc{MoEMA} generally improve over competing alternatives across simulations and real-data studies spanning tabular regression, conditional and inpainting image generation, and short-continuation text generation, showing the benefit of aggregating conditional generative models. \textsc{MoEMA} further improves on \textsc{StaticMA}. 

Two natural directions are left to future work. The first is to extend the method and the asymptotic theory beyond the i.i.d.\ setting, in particular to time-series data, where the independence assumptions underlying the present analysis no longer apply. The second is to extend both the methodology and the asymptotic theory to joint-distribution criteria and other IPM choices (see Subsections~H.1 and~H.2 of the Supplementary Material for details), which would broaden the class of distributional comparisons available for aggregation.

\bibliographystyle{abbrvnat}
\bibliography{bibliography}

\end{document}